\pdfoutput=1

\documentclass[11pt]{article}

\usepackage{acl}

\usepackage{times}
\usepackage{latexsym}

\usepackage[T1]{fontenc}

\usepackage[utf8]{inputenc}

\usepackage{microtype}

\usepackage{inconsolata}

\usepackage{graphicx}

\usepackage{amsmath}
\usepackage{xcolor}
\usepackage{tcolorbox}
\usepackage{tabularx,booktabs, array}
\newcolumntype{L}[1]{>{\raggedright\arraybackslash}p{#1}}
\usepackage{xurl} 
\usepackage{natbib}
\usepackage{caption}     
\usepackage{tabularx}    
\usepackage{booktabs}    
\usepackage{xltabular}   
\usepackage{array}       
\usepackage{xspace} 

\newtcbox{\langbox}{on line,
  colback=gray!10!white, colframe=gray!80!black,
  boxrule=0.3pt, arc=3pt, boxsep=1pt, left=1pt, right=1pt, top=1pt, bottom=1pt,
  fontupper=\ttfamily\footnotesize
}
\definecolor{palepeach}{RGB}{255,227,192}

\newcommand{\ie}{\textit{i.e., \xspace}}
\newcommand{\eg}{\textit{e.g., \xspace}}
\newcolumntype{L}[1]{>{\raggedright\arraybackslash}p{#1}}

\title{A Survey of Multilingual Reasoning  in Language Models}

\newcounter{eqfn} 
\makeatletter
\long\def\@makefntext#1{\parindent 1em\noindent
    \hb@xt@0.8em{\hss\@makefnmark\hspace{0.5em}}#1} 
\makeatother

\def\equalcontrib{%
  \ifnum\value{eqfn}=0
    \thanks{\*These authors contributed equally.}%
    \setcounter{eqfn}{\value{footnote}}
  \else
    \footnotemark[\value{eqfn}]%
  \fi%
}

\author{
    Akash Ghosh$^{1}$\thanks{Equal contribution. Work done while interning at Aikyam Lab (UVA). Contact author: \href{mailto:akash_2321cs19@iitp.ac.in}{akash\_2321cs19@iitp.ac.in}
    } \quad
    Debayan Dutta$^{1}$\footnotemark[1] \quad
    Sriparna Saha$^{1}$ \quad
    Chirag Agarwal$^{2}$ \\
    \\
    $^{1}$Indian Institute of Technology Patna, India \\
    $^{2}$University of Virginia, USA \\
}

\begin{document}
\maketitle
\begin{abstract}
While reasoning and multilingual capabilities in Language Models (LMs) have achieved remarkable progress in recent years, their integration into a unified paradigm—multilingual reasoning—is at a nascent stage. Multilingual reasoning requires language models to handle logical reasoning across languages while addressing misalignment, biases, and challenges in low-resource settings. This survey provides the first in-depth review of multilingual reasoning in LMs. In this survey, we provide a systematic overview of existing methods that leverage LMs for multilingual reasoning, specifically outlining the challenges, motivations, and foundational aspects of applying language models to reason across diverse languages. We provide an overview of the standard data resources used for training multilingual reasoning in LMs and the evaluation benchmarks employed to assess their multilingual capabilities. Next, we analyze various state-of-the-art methods and their performance on these benchmarks. Finally, we explore future research opportunities to improve multilingual reasoning in LMs, focusing on enhancing their ability to handle diverse languages and complex reasoning tasks. Rapid growth of evolving developments in this field can be actively tracked on our project page: \url{https://github.com/AkashGhosh/Survey-of-Multilingual-Reasoning-in-Language-Models}
\end{abstract}

\section{Introduction}

\label{sec:intro}
\begin{flushright}
    \normalsize
    \textit{If we spoke a different language, we would perceive a somewhat different world.} \vspace{-0.05in}

    \rule{0.9\linewidth}{0.4pt}  

    \vspace{-0.05in}
    \textit{~~~Ludwig Wittgenstein}
\end{flushright}


\looseness=-1 Large Language Models (LLMs)~\citep{vaswani2017attention} have emerged as transformative tools in natural language processing, demonstrating state-of-the-art performance in language generation, translation, and summarization\cite{jain2022survey,ghosh2024clipsyntel,ghosh2024healthalignsumm,ghosh2024medsumm,ghosh2025infogen,ghosal2025relic}. These models, trained on vast corpora, excel in generating human-like text and understanding diverse linguistic contexts. Despite their success in language generation, LLMs often face significant challenges in addressing \textit{underrepresented languages} and \textit{reasoning}. 

\looseness=-1 While the development of Multilingual LLMs ~\citep{qin2024multilingual,huang2024survey} extends LLM's capabilities in addressing multiple languages and catering to the needs of linguistically diverse communities, their proficiency in generation stems from training on large-scale corpora optimized for next-word prediction rather than logical inference~\citep{ramji2024inductive}. Consequently, while they produce fluent and contextually appropriate responses, they frequently struggle with complex reasoning tasks, particularly those requiring multi-step logic or nuanced understanding~\citep{patel2024multi}. These limitations become even more pronounced in multilingual settings due to key technical problems like cross-lingual misalignment, biases in training data, and the scarcity of resources for low-resource languages.

\looseness=-1 Reasoning is formally defined as the process of drawing logical conclusions, enabling individuals and systems to solve problems and make complex decisions. 
\looseness=-1 Recent advancements have sought to enhance the reasoning capabilities of LLMs using Chain-of-Thought (CoT)~\citep{wei2022chain}, fine-tuning~\citep{lobo2024impact}, and hybrid modeling~\citep{yao2024hdflow}, especially in high-resource languages like English. However, reasoning in multilingual contexts remains a relatively \textbf{unexplored} domain, where existing efforts predominantly focus on a handful of high-resource languages, leaving low-resource and typologically distant languages \textbf{underrepresented}. The lack of robust benchmarks, diverse training corpora, and alignment strategies further impede progress in this vital area.

\looseness=-1 Multilingual reasoning, which combines logical inference with multilingual capabilities, is essential for creating AI systems that effectively operate across diverse linguistic and cultural contexts~\citep{shi2022language}. Such systems hold immense potential for global applications, from multilingual education to culturally adaptive healthcare, ensuring inclusivity and fairness. The motivation for this survey arises from the urgent need to address these challenges and provide a systematic exploration of methods, resources, and future directions for multilingual reasoning in LLMs. The key contributions of our work are:
\begin{figure*}[htp]
    \centering
    \includegraphics[width=0.9\textwidth, height=10cm]{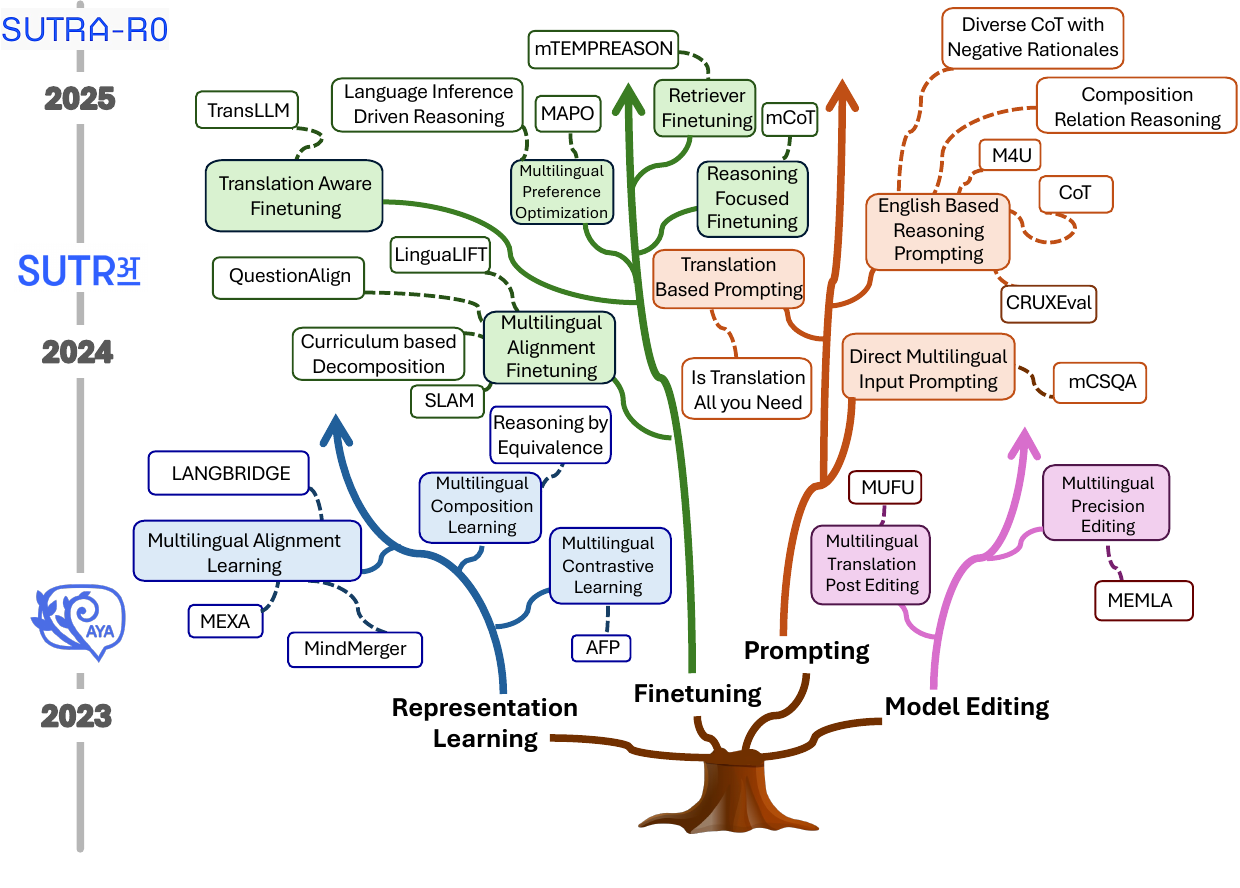} 
    \vspace{-0.25in}
    \caption{\textbf{Taxonomy tree of current Multilingual Reasoning Research.} The thrusts for improving multilingual reasoning mainly include representation learning, fine-tuning, prompting, and model editing. With the emergence of multilingual LLMs, while initial research focused on naive prompting, recent works propose several alignment, editing, and fine-tuning strategies to improve reasoning in multilingual LLMs.}
    \vspace{-0.1in}
    \label{fig:taxonomy_diagram}
\end{figure*}

\noindent\textbf{1) Comprehensive Overview:} We systematically review existing methods that leverage LLMs for multilingual reasoning, outlining challenges, motivations, and foundational aspects of applying reasoning to diverse languages.

\noindent\textbf{2) Training Corpora and Evaluation Benchmarks:} We analyze the strengths, limitations, and suitability of existing multilingual corpora and evaluation benchmarks in assessing the reasoning capabilities of LLMs for diverse linguistic tasks.

\noindent\textbf{3) Analysis of State-of-the-Art Methods:} We evaluate the performance of various state-of-the-art techniques, including CoT prompting, instruction tuning, and cross-lingual adaptations, on multilingual reasoning benchmark tasks.

\noindent\textbf{4) Future Research Directions:} We identify key challenges and provide actionable insights for advancing multilingual reasoning, focusing on adaptive alignment strategies, culturally aware benchmarks, and methods for low-resource languages.

\section{Multilingual Reasoning in LLMs}
\label{sec:multilingual-llms}

\looseness=-1 Recent advancements in LLMs have improved their reasoning capabilities. However, extending them across languages introduces several challenges, including consistency, low-resource adaptation, and cultural integration. Below, we describe the preliminaries and key characteristics of multilingual reasoning, focusing on challenges and desiderata for cross-lingual inference.

\subsection{Preliminaries}
\label{sec:prelim}

\looseness=-1\textbf{Large Language Models (LLMs).} LLMs are transformer-based neural network architectures designed to model the probability of a sequence of tokens. Formally, LLMs are trained to predict the likelihood of a word (or sub-word token) given the preceding words in a sequence \( X = \{x_1, \dots, x_n\} \), \ie $P(X) = \prod_{i=1}^n P(x_i \mid x_1, \dots, x_{i-1}),$ where $P(X)$ is the probability of the entire sequence and $P(x_i | x_1, \dots, x_{i-1})$ is the conditional probability of the i$^{th}$ token given the preceding tokens.


\noindent\textbf{Reasoning.} One of the key reasons behind the success of LLMs in mathematical and logical tasks is their reasoning capabilities. Formally, reasoning enables LLMs to draw logical conclusions \( C \) from premises \( P \) using a mapping function: $C = f(P)$. To this end, there are different types of reasoning strategies that an LLM can employ:\par
\noindent\textbf{a) Deductive Reasoning:} Derives logically certain conclusions from general premises. 
If the premises $P_i$ are true, the conclusion $C$ must also be true, \ie $P_1, P_2, \dots, P_n \;\Rightarrow\; C.$

\noindent\textbf{b) Inductive Reasoning:} Infers general rules or patterns from specific observations, 
leading to conclusions that are likely but not guaranteed, \ie $P_1, P_2, \dots, P_n \;\Rightarrow\; C_{\text{probabilistic}}.$

\noindent\textbf{c) Abductive Reasoning:} Infers the most plausible hypothesis ($H_{\text{best}}$) that explains an observation $O$, 
though the inference is not guaranteed to be correct, \ie $O \;\Rightarrow\; H_{\text{best}}.$

\noindent\textbf{d) Analogical Reasoning:} Transfers knowledge by identifying relational similarities between domains, 
\ie $A : B \;\approx\; C : D.$

\noindent\textbf{e) Commonsense Reasoning:} Draws on background knowledge of everyday situations to make intuitive, 
contextually appropriate inferences.

\subsection{Desiderata in Multilingual Reasoning}
\label{sec:desiderata} 
Here, we describe desiderata that lay the foundation for multilingual reasoning in LLMs. 
Let \( L {=} \{l_1, l_2, \dots, l_m\} \) represent a set of \( m \) languages, and let \( P_l \) and \( C_l \) denote the premise and conclusion in a given language \( l_i \). For a multilingual reasoning model \( M \), the task can be defined as: $M(P_{l_{i}}) \to C_{l_{i}}, \quad \forall l_{i} \in L,$
where \( M \) must satisfy the following key desiderata:

\noindent\textbf{1. Consistency:} A model should make logically equivalent conclusions across languages for semantically equivalent premises, \ie $C_{l_i} \approx C_{l_j}, \quad \text{if } P_{l_i} \equiv P_{l_j}, \quad \forall l_i, l_j \in L,$
where \( \equiv \) indicates semantic equivalence of premises across languages. Consistency ensures that logical conclusions remain invariant of the input language.

\noindent\textbf{2. Adaptability:} For languages \( l_k \in L_{\text{low-resource}} \), the model must generalize effectively using cross-lingual transfer from high-resource languages and perform robust reasoning, \ie $\forall l_k \in L_{\text{low-resource}}, \quad M(P_{l_k}) \to C_{l_k}.$

\noindent\textbf{3. Cultural Contextualization:} Reasoning should consider cultural and contextual differences inherent to each language, \ie for a context $c_{l_i}$ specific to language $l_i$, the conclusion $C_{l_i}$ should adapt accordingly: $C_{l_i} = f(P_{l_i}, c_{l_i}), \quad \forall l_i \in L,$ where $f$ is a mapping function that integrates linguistic reasoning with cultural nuances.

\looseness=-1\noindent\textbf{4. Cross-Lingual Alignment:} 
The model must align reasoning processes across typologically diverse languages, where typology refers to linguistic differences in syntax, morphology, and structure (\eg word order variations between English and Japanese). Given the typological variations $T_{l_{i}}$ and $T_{l_{j}}$ for languages $l_{i}$ and $l_{j}$, alignment ensures that reasoning remains consistent and coherent across languages, \ie
$
\text{if } P_{l_i} \equiv P_{l_j}, \quad M(P_{l_i}) \approx M(P_{l_j}), \quad \forall l_i, l_j \in L.
$

\noindent Next, we highlight existing works that propose different training corpora and benchmarks for multilingual reasoning in Sec.~\ref{sec:dataset} and then describe previously proposed techniques to improve the multilingual reasoning of LLMs in Sec.~\ref{sec:method}. 

\section{Multilingual Reasoning Datasets}
\label{sec:dataset}
Models trained on english corpora exhibit language biases~\citep{lyu2024regional}, limiting their reasoning capability on non-English languages. Training an LM to solve math problems across languages requires multilingual understanding and mathematical reasoning~\citep{son2024mm}. Hence, multilingual datasets and benchmarks play a key role in training multilingual LMs and evaluating the effectiveness of various LMs and techniques in handling domain-specific reasoning queries across low- and high-resource languages~\citep{xu2024cruxeval, rasiah2024one,xue2024famma}. Below, we detail training datasets~(Sec.~\ref{sec:corpus}) and benchmarks~(Sec.~\ref{sec:eval-bench}), comprising domains, tasks, and language distribution in current multilingual reasoning datasets.

\subsection{Training Corpus}
\label{sec:corpus}
\looseness=-1 The best strategy to equip an LM with a specific type of reasoning is to train the model on it. However, the training objective differs based on the use case, domain, and language in which the model needs to be adapted. For example, to perform mathematical reasoning~\citep{cobbe2021training,amini2019mathqa} in a particular language, it needs to be trained with mathematical reasoning datasets, which will differ if we want to adapt the model for legal reasoning. 
\begin{figure*}
    \centering
    \includegraphics[width=\linewidth]{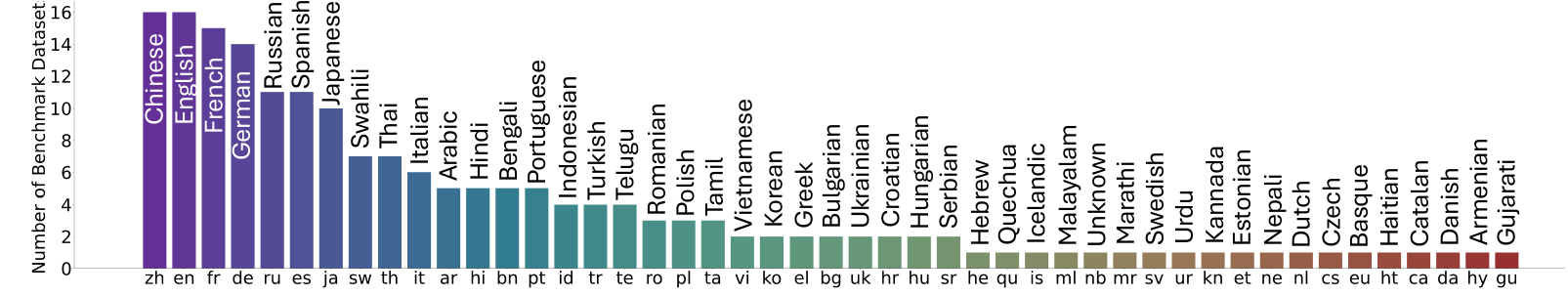}
    \vspace{-0.25in}
    \caption{\looseness=-1\textbf{Language distribution across training corpora and benchmarks for multilingual reasoning.} The \texttt{y}-axis denotes the number of training corpora/benchmark datasets that include a given language (\texttt{x}-axis). We observe a long-tail distribution, denoting that current datasets predominantly cover languages like Chinese, English, French, and German, highlighting the need for benchmarks that represent long-tail languages.}
    \vspace{-0.2in}
    \label{fig:four_plots}
\end{figure*}

\looseness=-1 While most training corpora are predominantly based on mathematical reasoning, XCSQA~\citep{zhu2024power} and MultiNLI~\citep{williams2017broad} are used for enhancing logical and coding reasoning, and sPhinX~\citep{ahuja2024sphinx} is developed to translate instruction-response pairs into 50 languages for fine-tuning. In addition, there are cases where translation datasets like OPUS~\citep{tiedemann2012opus}, FLORES-200~\citep{goyal2022flores}, and LegoMT~\citep{yuan2022legomt} are used to map the multilingual representation into the LM's representation space.
Further,~\citet{ponti2020xcopa} introduced XCOPA to show that multilingual pre-training and zero-shot fine-tuning underperform compared to translation-based transfer. We argue that, moving forward, selecting the appropriate dataset and training methodology is crucial for optimizing a model's performance in specialized reasoning tasks.

\subsection{Evaluation Benchmark}
\label{sec:eval-bench}
\looseness=-1 Benchmarks are key to advancing the field of multilingual reasoning as they provide a systematic framework to assess the performance of models across diverse reasoning tasks. Each reasoning task and domain presents unique challenges, making it crucial to have tailored benchmarks that reflect specific requirements and complexities of those tasks. Below, we analyze the evaluation benchmarks on three key aspects, namely languages~(Fig.~\ref{fig:four_plots}), domain~(Fig.~\ref{fig:pie-chart-1}), and task~(Fig.~\ref{fig:pie-chart-2}).

\subsubsection{Domains and Tasks Covered}
\label{sec:domain}
\begin{figure}[t]
    \centering
    \includegraphics[width=0.9\linewidth]{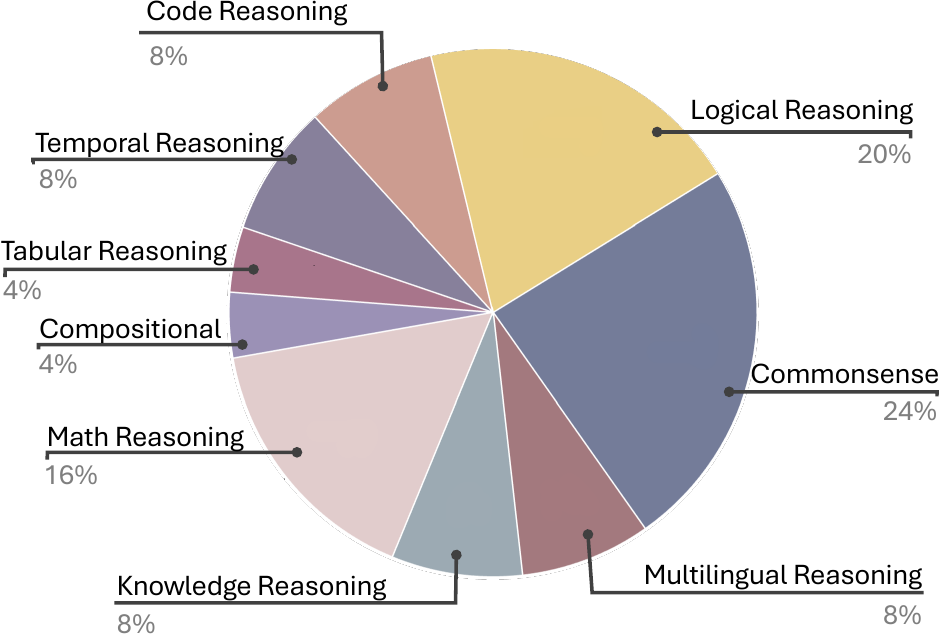}
    \vspace{-0.05in}
    \caption{\textbf{Distribution of multilingual reasoning datasets.} We find that datasets predominantly comprise logical, commonsense, and math reasoning, and the community needs benchmarks to include compositional and tabular reasoning.}
    \vspace{-0.125in}
    \label{fig:pie-chart-1}
\end{figure}
\begin{figure}[ht]
    \centering
    \includegraphics[width=0.9\linewidth]{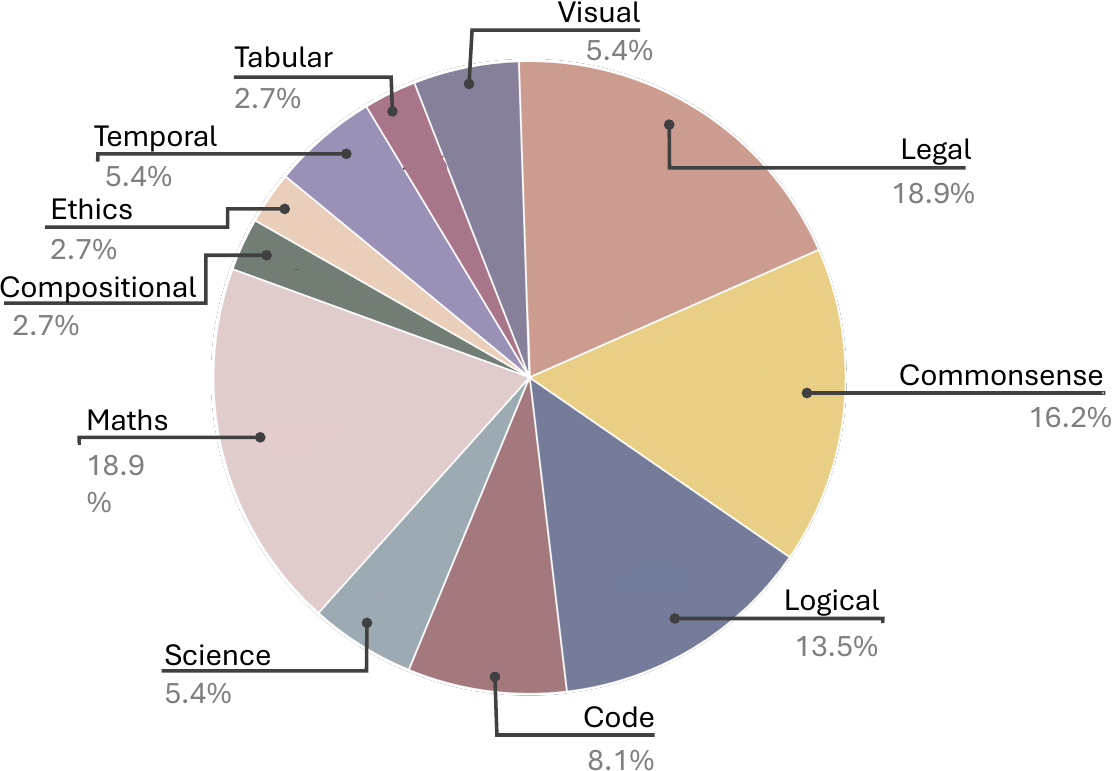}
    \vspace{-0.05in}
    \caption{\looseness=-1\textbf{Distribution of domains in multilingual reasoning datasets.} While legal, commonsense, and math domain dataset cover up to 54\% of current multilingual reasoning research, other under-explored domains include ethics, science, visual, and compositional.}
    \vspace{-0.2in}
    \label{fig:pie-chart-2}
\end{figure}
Multilingual reasoning in LMs spans multiple domains, each with its complexities and requirements, and understanding these differences is essential for developing LMs that can effectively adapt to various applications. For instance,~\citet{cobbe2021training} highlighted that mathematical reasoning requires structured multi-step logic and datasets. While~\citet{ponti2020xcopa} showed that causal reasoning in XCOPA relies on cross-lingual consistency and commonsense inference,~\citet{ostling2016continuous} noted that multilingual reasoning introduces typological challenges. 
These studies emphasize the need for tailored approaches to address the specific demands of each task and domain. Hence, it is crucial to \textbf{build reliable and robust benchmarks} for developing more robust techniques tailored to handle the complexity of a particular domain and task. Figs.~\ref{fig:pie-chart-1}-\ref{fig:pie-chart-2} show the distribution of datasets across various domains and tasks, highlighting the need to develop more comprehensive benchmarks across multiple domains. Currently, tasks such as math, legal, and commonsense reasoning dominate multilingual benchmarks, collectively accounting for \textbf{54\%} of the total (Fig.~\ref{fig:pie-chart-2}). In contrast, domains like science, ethics, and visual, tabular, and temporal reasoning are underrepresented, covering only \textbf{35\%}. Notably, \textbf{crucial domains such as finance and healthcare still lack dedicated evaluation benchmarks} for multilingual reasoning, highlighting a significant gap in the field.

\subsubsection{Languages Covered}
\label{sec:language}

\begin{figure*}
    \centering
\includegraphics[width=0.9\textwidth]{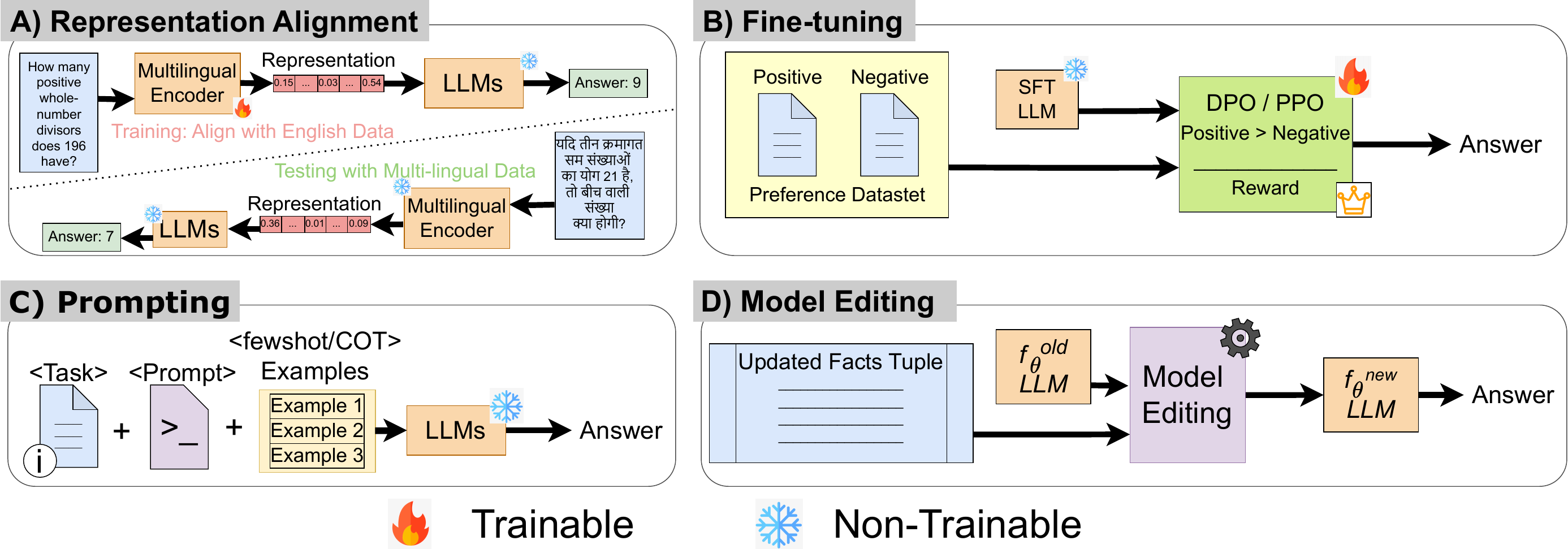}
    \vspace{-0.1in}
    \caption{\looseness=-1\textbf{Taxonomy of Multilingual Reasoning Methods.} A taxonomy of approaches for enhancing multilingual reasoning in models, covering (A) Representation Alignment, (B) Finetuning, (C) Prompting, and (D) Model Editing.}
    \vspace{-0.15in}
    \label{fig:method}
\end{figure*}

Comprehensive language coverage is vital for multilingual reasoning, ensuring inclusivity and balanced performance across low- and high-resource linguistic communities. Based on languages, current benchmarks can be primarily classified into human and coding languages. Benchmarks like XNLI~\citep{conneau2018xnli}, mCSQA~\citep{sakai2024mcsqa}, and m-ARC~\citep{lai2023okapi} predominantly focus on high-resource languages like English, Chinese, French, and Spanish. While some efforts include low-resource languages like Swahili (XCOPA~\citep{ponti2020xcopa}), Haitian (M4U~\citep{wang2024m4u}), and Nepali (mMMLU~\citep{hendrycks2020measuring}), \textbf{their representation remains minimal} and research in these languages remains at a nascent stage. Typologically distant and underrepresented languages, such as Kannada, Gujarati (xSTREET~\citep{li2024eliciting}), and Quechua, are rarely included, \textbf{further widening linguistic inequalities.} Datasets like FLORES-200 attempt to balance low- and high-resource languages but fail to achieve comprehensive coverage. To ensure effective LLM performance across diverse linguistic and cultural contexts, it is critical to include a broader range of low-resource and endangered languages~\citep{goyal2022flores, amini2019mathqa} (see the complete distribution of human languages across benchmarks in Fig.~\ref{fig:four_plots}). Finally, only four benchmarks~\citep{luo2024python,xu2024cruxeval,zhang2024p,li2024eliciting} incorporate coding languages across multiple languages.

\section{Methods}
\label{sec:method}
\looseness=-1 Multilingual reasoning within LMs has garnered significant attention in recent years, leading to the development of diverse techniques for enhancing their capabilities across diverse languages. Prior works have explored various directions to improve multilingual reasoning. Building upon this body of work (see Fig.~\ref{fig:method}), we identify four primary thrusts, \textit{viz.} representation alignment, fine-tuning, prompting, and model editing, collectively contributing to advancing multilingual reasoning in LMs.\\
\looseness=-1\noindent\textbf{a) Representation Alignment.} Multilingual reasoning requires consistent representations across languages, but LMs often struggle due to imbalanced training data. Representation alignment ensures that equivalent concepts share similar embeddings, reducing inconsistencies in multilingual inference, vital for reasoning and multilingual generalization. \citet{li2024improving} employs contrastive learning to align multilingual sentence representations by treating translation pairs as positive samples and pulling their embeddings closer, bridging language representation gaps and enhancing model's cross-lingual reasoning and generation capabilities. Multilingual Alignment Learning is another technique that ensures semantic consistency across languages by aligning their representations for improved multilingual performance~\citep{huang2024mindmerger}, bridging multilingual encoders with LLMs using minimal parameters to achieve effective alignment without supervision~\citep{yoon2024langbridge,kargaran2024mexa}. Similarly,~\citet{ruan2025layalign} integrates all encoder layer representations and employs adaptive fusion-enhanced attention to enable layer-wise alignment between the LLM and multilingual encoder, ensuring consistent cross-lingual representations and improving the model's multilingual reasoning capabilities. Finally, an exciting new direction is multilingual compositional learning, which constructs compositional representations by combining equivalent token embeddings across multiple languages~\citep{arora2024towards} and formalizing problems in an abstract space and solving them step-by-step using self-training for improved alignment across languages~\citep{ranaldi2025multilingual}.

\looseness=-1\noindent\textbf{b) Finetuning.} It leverages cross-lingual data and tasks to fine-tune models for enhanced reasoning and comprehension, leading to numerous innovative approaches. For instance, LinguaLIFT~\citep{zhang2024lingualift} uses code-switched fine-tuning along with language alignment layers to effectively bridge the gap between English and low-resource languages, helping maintain the nuance and context across linguistic boundaries. Similarly, QuestionAlign~\citep{zhu2024power} aligns questions and responses in multiple languages, thereby enhancing cross-lingual understanding and consistency in reasoning and~\citet{ko2025understand} introduces a strategic fine-tuning approach that anchors reasoning in English and then translates results, significantly reducing cross-lingual performance gaps. Strategic fine-tuning using a small but high-quality bilingual dataset can enhance both the reasoning capabilities and non-English language proficiency of LLMs~\citep{ha2025pensez}. While these methods have leaned towards extensive fine-tuning, SLAM~\citep{fan2025slam} introduces a more parameter-efficient strategy and selectively tunes layers critical for multilingual comprehension, \textbf{significantly lowering the computational demands} while still maintaining or even enhancing the model's reasoning capabilities. Translation has also been harnessed as a powerful tool for knowledge transfer in multilingual settings, where TransLLM~\citep{geng2024not} focuses on translation-aware fine-tuning to align different languages, enhancing language understanding but also adapting the model for various cross-lingual tasks. For those aiming at more complex reasoning tasks, \textbf{reasoning-focused fine-tuning} has proven beneficial. The Multilingual CoT (mCoT) instruction tuning method~\citep{lai2024mcot} utilizes a dataset specifically curated for reasoning across languages and combines CoT reasoning with instruction tuning to boost consistency and logical problem-solving in multiple languages. In addition, preference-based techniques to align reasoning outputs across languages emphasize the use of language imbalance as a reward signal in models like Direct Preference and Proximal Policy Optimization~\citep{she2024mapo}. Recent research has demonstrated that Process Reward Modeling offers fine-grained feedback at each step of the reasoning process, only~\citet{wang2025demystifying} has shown its application on non-English language. Finally, an interesting direction moving forward is curriculum-based and retriever-based fine-tuning techniques to enhance multilingual reasoning~\citep{anand2024multilingual,bajpai2024multilingual}, where models must not only retrieve relevant information but also compare them to evaluate relationships between them~\citep{agrawal2024evaluating,ranaldi2025improving,shao2024deepseekmathpushinglimitsmathematical,yang2025mr}.

\noindent\textbf{c) Prompting.} 
\looseness=-1 Prompting has emerged as a key technique for enhancing how LLMs adapt and reason across different languages. By guiding the model through specific strategies, prompting facilitates dynamic language adaptation and addresses the data imbalance challenge, thereby enhancing cross-lingual consistency, logical alignment, and the robustness of reasoning. For instance, an effective method is Direct Multilingual Input Prompting~\citep{sakai2024mcsqa}, where the model directly processes inputs in various native languages without translation, preserving the original linguistic nuances. This approach was notably applied in the paper ``\textit{Do Moral Judgements}''~\citep{khandelwal2024moral}, where moral scenarios were directly presented in their native languages to assess the model's reasoning capabilities. Another strategy, Translation-based prompting~\citep{liu2024translation} uses translation to convert multilingual inputs into a target language for processing, where tasks are translated into English for reasoning and translated back to the target language for evaluation~\citep{wang2024m4u,zhao2024large}. This is also used to generate diverse CoT with Negative Rationales by incorporating both correct and incorrect reasoning paths to refine multilingual reasoning capabilities~\citep{payoungkhamdee2024empirical}. 
While in-context learning with natural language can be ambiguous and less effective in low-resource languages, program-based demonstrations offer clearer, structured reasoning that transfers better across languages~\citep{ranaldi2025natural}. In addition to the above strategies, Dictionary Insertion Prompting (DIP) offers a lightweight and practical alternative by inserting English translations of keywords into non-English prompts, bridging linguistic gaps without full translation and enabling clearer reasoning and improved performance in multilingual tasks~\citep{lu2024dictionary}.

\begin{figure*}[h]
    \centering
    \includegraphics[width=0.63\textwidth]{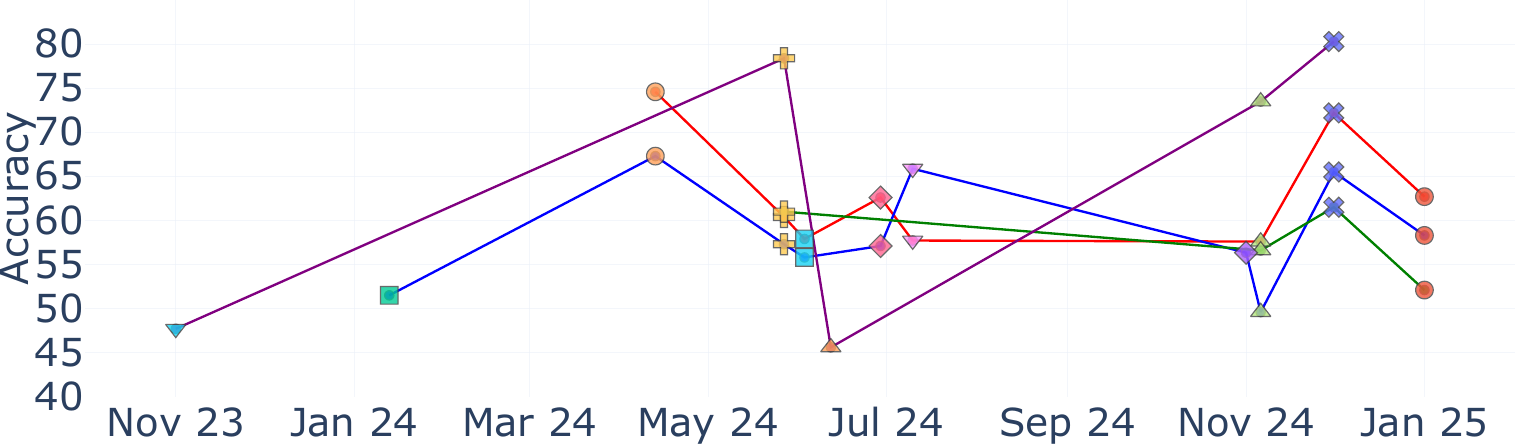}
    \caption{\looseness=-1 \small Accuracy trends of various methods on multilingual reasoning benchmarks, including \colorbox{blue!40}{MGSM}, \colorbox{red!40}{MSVAMP}, \colorbox{purple!40}{XNLI}, and \colorbox{green!40}{XCSQA}. The $x$-axis represents the arXiv paper submission date, and the $y$-axis indicates percentage accuracy.}
    \label{fig:benchmark}
\end{figure*}

\noindent\textbf{d) Model Editing.} 
\looseness=-1  Model editing is a growing and exciting research area that aims to modify/update the information stored in a model. Formally, model editing strategies update pre-trained models for specific input-output pairs without retraining them and impacting the baseline model performance on other inputs. Multilingual Precision Editing involves making updates to model knowledge while ensuring minimal impact on unrelated information. Multilingual knowledge Editing with neuron-Masked Low-Rank Adaptation (MEMLA)~\citep{xie2024memla} enhances multilingual reasoning by leveraging neuron-masked LoRA-based edits to \textbf{integrate knowledge across languages and improve multi-hop reasoning} capabilities. Further, Multilingual Translation Post-editing refines translations by correcting errors in multilingual outputs for better alignment, where we can enhance multilingual reasoning by incorporating auxiliary translations into the post-editing process, enabling LLMs to improve semantic alignment and translation quality across languages~\citep{lim2024mufu}.

\looseness=-1 \noindent\textbf{An emerging complementary direction} investigates inference-time (test-time) compute scaling in enhancing multilingual reasoning. Recent work shows that scaling up compute for English-centric reasoning language models (RLMs) can significantly improve performance across many languages, including low-resource ones, even surpassing larger models~\citep{yong2025crosslingual}. While most test-time techniques, such as CoT prompting with trial and error, have primarily focused on English, methods like English-Pivoted CoT training~\citep{tran2025scaling} exploit the model's strong English reasoning capabilities to support multilingual tasks, offering a promising path to bridge alignment gaps for underrepresented languages.

\section{Evaluation Metrics and Benchmarks}
\looseness=-1 Evaluating multilingual reasoning in LLMs requires standardized metrics to ensure logical consistency and cross-lingual coherence. Unlike traditional NLP, it must address inference errors, translation drift, and reasoning stability across languages.

\subsection{Metrics}
\looseness=-1 Here, we detail key metrics for evaluating multilingual reasoning, along with their formal definitions:

\looseness=-1\noindent\textbf{1) Accuracy.} These metrics assess overall correctness in reasoning and multilingual benchmarks: i) \textit{General Accuracy} measures the proportion of correct outputs over total samples, and ii) \textit{Zero-Shot Accuracy}, which evaluates model performance on unseen tasks or categories without fine-tuning.

\noindent\textbf{2) Reasoning and Consistency.} These metrics evaluate logical inference and multi-step reasoning ability: i) \textit{Reasoning Accuracy} assesses correctness in logical and step-by-step reasoning tasks and ii) \textit{Path Consistency} measures coherence between reasoning steps in CoT prompting.

\looseness=-1\noindent\textbf{3) Translation and Cross-Lingual.} To ensure multilingual reasoning consistency, models must preserve meaning across languages: i) \textit{Translation Success Rate} measures correctness and semantic preservation in multilingual translations as the ratio of accurate translations and total translations and ii) \textit{Cross-Lingual Consistency} evaluates whether logically equivalent statements yield \textit{consistent reasoning outputs} across different languages.

\noindent\textbf{4) Perplexity and Alignment.} They quantify \textit{semantic alignment} and measure whether embeddings across languages remain consistent: i) \textit{Perplexity-Based Alignment ($P_{\text{align}}$)}
\begin{equation}
    P_{\text{align}} = \exp\left(-\frac{1}{N} \sum_{i=1}^{N} \log P(x_i)\right),
\end{equation}
\looseness=-1 where $P(x_i)$ is the model's probability of predicting token $x_i$ (lower perplexity means better alignment) and ii) \textit{Semantic Alignment} measures the cosine similarity between multilingual sentence embeddings: $S_{\text{align}} = \frac{E_l \cdot E_t}{\|E_l\| \|E_t\|},$ where $E_l$ and $E_t$ are sentence embeddings in different languages.

\subsection{Performance on Benchmarks}
\looseness=-1 Here, we discuss the performance of the aforementioned methods on standard mathematical (MGSM~\citep{shi2022language}, MSVAMP~\citep{chen2023breaking}), commonsense (xCSQA~\citep{lin2021xcsqa}), logical (xNLI~\citep{conneau2018xnli}) reasoning benchmarks\footnote{We only cover benchmarks analyzed by more than four papers.}. Next, we describe the four most popular benchmarks and detail the performance of reasoning techniques, highlighting existing model gaps that limit their reasoning performance.

\looseness=-1\noindent\textbf{MGSM} tests multilingual arithmetic reasoning in LMs with 250 translated math problems in ten diverse languages. 
While recent trends suggest that advanced post-training techniques like MAPO are key for strong performance, fine-tuning strategies may be more impactful than stronger reasoning architectures or relying on the model's English expertise to improve multilingual performance.

\looseness=-1\noindent\textbf{MSVAMP} is an out-of-domain multilingual mathematical reasoning dataset comprising 10k problems across ten languages and serves as a comprehensive test bed to evaluate LMs' generalization in multilingual mathematical contexts. We find that advanced preference optimization achieves much stronger performance than CoT-based fine-tuning, suggesting advanced fine-tuning techniques are a better direction to beat the current best in this benchmark.

\looseness=-1\noindent\textbf{xCSQA} is a multilingual extension of the CommonsenseQA dataset, encompassing 12,247 multiple-choice questions translated into 15 languages, designed to assess LMs' cross-lingual commonsense reasoning capabilities. The current trend shows that stronger fine-tuning strategies like two-step fine-tuning or preference optimization show better performance than selectively fine-tuning specific layers as in SLAM.

\looseness=-1\noindent\textbf{xNLI} evaluates cross-lingual inference across 15 languages. Recent studies suggest that LM integration with external models~\citep{huang2024mindmerger} and multilingual alignment followed by fine-tuning~\citep{zhang2024lingualift} outperform contrastive learning methods like TCC~\citep{chia2023contrastive}, highlighting the need for more structured multilingual adaptation strategies.

\section{Future Directions}
\label{sec:future}
\looseness=-1 With the rapid development of reasoning models, our community must ensure that models remain unbiased towards low-resource languages. Looking forward, we call on the community to put their collective efforts into the following directions:

\looseness=-1\noindent\textbf{1. Multilingual Alignment and Reasoning Transfer.} A key challenge in multilingual reasoning is the lack of data in different languages. One promising solution is to leverage existing large datasets and transfer/distill their knowledge in the representation space~\citep{yoon2024langbridge,huang2024mindmerger}. Future research should develop cross-lingual knowledge transfer techniques, enabling models to use high-resource languages as a bridge to enhance reasoning in \textit{low-resource languages}. Another direction is to generate synthetic datasets using techniques like back-translation and data augmentation, tailored specifically for reasoning tasks.

\looseness=-1\noindent\textbf{2. Explainable and Interpretable Reasoning.} Ensuring faithful reasoning in multilingual LLMs is challenging due to linguistic diversity, translation ambiguities, and reasoning inconsistencies. Studies on English CoT reasoning ~\citep{tanneru2024hardness,lobo2024impact} highlight faithfulness issues, which become more severe when extended to low-resource languages. Causal reasoning can enhance cross-lingual alignment, improving interpretability by uncovering cause-and-effect relationships across languages. Future research should focus on integrating causal reasoning and multilingual CoT frameworks to ensure logical coherence, transparency, and trust in multilingual AI systems.

\looseness=-1\noindent\textbf{3. Advanced Training and Inference Techniques.} While recent advancements in multilingual reasoning have introduced reasoning-aware fine-tuning and multilingual preference optimization techniques, further efforts are needed to improve training paradigms. Some exciting techniques in this direction includes post-training RL methods that improve reasoning in low-resource languages~\citep{wu2024reuse} and efficient inference-time scaling and Agentic frameworks~\citep{khanov2024args,chakraborty2024transfer}. Preliminary post-training works~\citep{xuan2025mmlu} show that they yield mixed results across languages, with effectiveness depending on the base model and required degree of linguistic diversity, highlighting the need for language inclusive training approaches.

\looseness=-1\noindent\textbf{4. Unified Evaluation Metrics.} A comprehensive evaluation framework is a crucial missing component for assessing multilingual reasoning capabilities. Metrics should measure logical consistency, cultural adaptability, and robustness, considering real-world and adversarial multilingual settings.

\looseness=-1\noindent\textbf{5. Multimodal Multilingual Reasoning.} While there are a few works on visual reasoning in the multilingual context~\citep{das2024exams,gao2025pm4bench,ghosh2024exploring}, multimodal reasoning (integrating tables, text, image, audio, and video) remains largely unexplored. Advancing this area could enable models to handle complex tasks in low-resource languages and incorporate cross-modal reasoning.

\looseness=-1\noindent\textbf{6.New Benchmarks:} As multilingual reasoning advances, robust evaluation benchmarks are essential because reasoning is highly domain-specific in nature, developing targeted benchmarks is crucial, especially in high-stakes fields like healthcare, law, and finance, where accuracy directly affects decision-making. For instance,~\citet{xue2024famma} introduces FAMMA which shows significant challenges in the field of Financial Question Answering.\par

\looseness=-1\noindent\textbf{7. Efficient Reasoning Models.} An emerging direction in reasoning research is enhancing resource efficiency in reasoning-aware models. Recent works like \cite{ning2024can} propose strategies for more efficient reasoning, reducing computational costs while maintaining logical consistency. However, this area remains largely unexplored in multilingual settings, offering a key opportunity to develop scalable reasoning models that generalize across languages with minimal resources.\par

\looseness=-1\noindent\textbf{8. Miscellaneous Tasks.} LLMs have achieved remarkable performance across a wide range of tasks; however, they continue to struggle with complex compositional reasoning~\citep{zhao2024exploring}, often performing only marginally better than random guessing. They also face difficulties in reasoning over longer contexts, particularly in low-resource languages~\citep{hengle2025can}. Moreover, their reasoning traces frequently exhibit hallucinations~\citep{sahoo2024unveiling}, with models failing to reliably integrate information or recognize missing pieces even when the relevant facts are retrievable.

\section{Conclusion}
\looseness=-1 Multilingual reasoning in LLMs is a rapidly evolving field, addressing critical challenges like cross-lingual alignment, low-resource language gaps, and cultural adaptation. Our survey highlights advancements in fine-tuning, prompting, and representation learning while identifying gaps in scalability and domain-specific applications. It serves as a call to action for the LLM and reasoning community to focus on advanced alignment techniques, culturally aware reasoning, and scalable architectures. By breaking language barriers and fostering inclusivity, multilingual reasoning can create globally impactful AI systems. Our survey provides a foundation for advancing research in this transformative domain.

\section{Limitations}
 This is the first survey dedicated to the important and emerging topic of multilingual reasoning. We have made every effort to include key studies and recent advancements in this area; however, we acknowledge that some relevant work may have been unintentionally missed. As the field is still in its early stages, this survey does not aim to provide definitive solutions for improving multilingual reasoning. Instead, our goal is to analyze existing approaches and offer a comprehensive evaluation of which techniques demonstrate stronger performance across current benchmarks.

\section{Acknowledgement}
We would like to thank the anonymous reviewers for their insightful feedback. C.A. is supported, in part, by grants from Capital One, LaCross Institute for Ethical AI in Business, the UVA Environmental Institute, OpenAI Researcher Program, and Cohere. The views expressed are those of the authors and do not reflect the official policy or position of the funding agencies.

\bibliography{latex/acl_latex}

\newpage

\appendix

\onecolumn
\section{Appendix}
\label{sec:appendix}

\noindent\textbf{Related Surveys}
The earliest surveys \cite{qin2024multilingual,xu2025survey}—both from April 2024 focus on laying foundational taxonomies of Multilingual LLMs(MLLMs):\cite{qin2024multilingual} survey resources, taxonomy, and emerging frontiers in MLLMs, while \cite{xu2025survey} delves deeply into multilingual corpora, alignment techniques, and bias issues. \citet{huang2024survey} broadens the scope to multiple perspectives—training/inference, security, cultural domains, and datasets—framing “new frontiers” in multilingual LLM research. Finally, the survey by  \cite{zhu2024multilingual} provides the most comprehensive “systematic” treatment: it covers architectures, pre-training objectives, alignment datasets, a detailed evaluation roadmap (including safety, interpretability, reasoning), and real-world applications across domains. While there are a lot of surveys on general reasoning in LLMs \cite{bandyopadhyay2025thinking,chen2025towards},\textit{this survey is the first survey dedicated specifically to multilingual reasoning, drilling deeply into logical inference across languages, its unique challenges (misalignment, bias, low-resource gaps), and the benchmarks and methods tailored to evaluate and improve reasoning capabilities.}

\noindent\textbf{Distribution of languages in Reasoning Datasets.}

\noindent We show a detailed tabular format of the languages used in different reasoning datasets along with their languages.

\begin{table}[ht]
\centering
    \renewcommand{\arraystretch}{0.9}
    \setlength{\tabcolsep}{3pt}  
\begin{tabular}{
    l l @{\hspace{3em}} 
    l l @{\hspace{3em}} 
    l l @{\hspace{3em}} 
    l l
}
\toprule
\langbox{af} & Afrikaans & \langbox{ar} & Arabic     & \langbox{be} & Belarusian & \langbox{bg} & Bulgarian \\
\langbox{bn} & Bengali   & \langbox{ca} & Catalan    & \langbox{cs} & Czech      & \langbox{da} & Danish \\
\langbox{de} & German    & \langbox{el} & Greek      & \langbox{en} & English    & \langbox{es} & Spanish \\
\langbox{et} & Estonian  & \langbox{eu} & Basque     & \langbox{fa} & Persian    & \langbox{fi} & Finnish \\
\langbox{fr} & French    & \langbox{ha} & Hausa      & \langbox{he} & Hebrew     & \langbox{hi} & Hindi \\
\langbox{hr} & Croatian  & \langbox{ht} & Haitian    & \langbox{hu} & Hungarian  & \langbox{hy} & Armenian \\
\langbox{id} & Indonesian & \langbox{id} & Indonesian & \langbox{is} & Icelandic & \langbox{it} & Italian \\
\langbox{ja} & Japanese  & \langbox{kn} & Kannada    & \langbox{ko} & Korean     & \langbox{lb} & Luxembourgish \\
\langbox{mk} & Macedonian & \langbox{ml} & Malayalam & \langbox{mr} & Marathi    & \langbox{nb} & Norwegian Bokmal \\
\langbox{ne} & Nepali    & \langbox{nl} & Dutch      & \langbox{pl} & Polish     & \langbox{pt} & Portuguese \\
\langbox{qu} & Quechua   & \langbox{ro} & Romanian   & \langbox{ru} & Russian    & \langbox{sk} & Slovak \\
\langbox{sl} & Slovenian & \langbox{sr} & Serbian    & \langbox{sv} & Swedish    & \langbox{tr} & Turkish \\
\langbox{uk} & Ukrainian & \langbox{ur} & Urdu       & \langbox{vi} & Vietnamese & \langbox{zh} & Chinese \\
\bottomrule
\end{tabular}
\caption{Language Codes and Their Corresponding Languages}
\end{table}

\renewcommand{\arraystretch}{1.3}
\begin{table}[h]
\centering
\small
\renewcommand{\arraystretch}{0.9}
\setlength{\tabcolsep}{1pt}  
\captionof{table}{Multilingual Datasets and their respective papers, domains, and languages.}
\begin{tabularx}{\textwidth}{L{3cm} L{6cm} L{3cm} L{3cm}}
\toprule
\textbf{Dataset} & \textbf{Paper} & \textbf{Domain} & \textbf{Languages} \\
\midrule
MSVAMP & \citep{she2024mapo,yoon2024langbridge,zhu2024question,zhu2024power,lai2024mcot,chai2401xcot,huang2024mindmerger,zhang2024lingualift,fan2025slam} & Maths & \langbox{zh}, \langbox{th}, \langbox{ja}, \langbox{en}, \langbox{de}, \langbox{fr}, \langbox{es}, \langbox{bn.} \langbox{sw} \\ \midrule
MGSM & \citep{she2024mapo,yoon2024langbridge,zhu2024question,zhu2024power,lai2024mcot,chai2401xcot,huang2024mindmerger,liu2024translation,zhang2024lingualift,fan2025slam} & Maths & \langbox{zh}, \langbox{th}, \langbox{ja}, \langbox{en}, \langbox{de}, \langbox{fr}, \langbox{es}, \langbox{ru}, \langbox{bn.} \langbox{sw}, \langbox{te} \\ \midrule
MNumGLUESub & \citep{she2024mapo} & Maths & \langbox{bn}, \langbox{th}, \langbox{sw}, \langbox{ja}, \langbox{zh}, \langbox{ru}, \langbox{de}, \langbox{es}, \langbox{fr}, \langbox{en} \\ \midrule
MetaMathQA & \citep{yoon2024langbridge,zhu2024question,zhu2024power,lai2024mcot,huang2024mindmerger} & Maths & \langbox{en} \\ \midrule
Proof-Pile 2 & \citep{yoon2024langbridge} & Maths & \langbox{en} \\ \midrule
Exams Dataset & \citep{payoungkhamdee2024empirical} & Science and Humanities & \langbox{ar}, \langbox{de}, \langbox{fr}, \langbox{es}, \langbox{it}, \langbox{pl}, \langbox{vi}, \langbox{pt}, \langbox{sr}, \langbox{hu}, \langbox{tr}, \langbox{bg}, \langbox{hr}, \langbox{mk}, \langbox{sq} \\ \midrule
M4U Benchmark & \citep{wang2024m4u} & Science & \langbox{zh}, \langbox{en}, \langbox{de} \\ \midrule
XCSQA & \citep{zhu2024power,zhang2024lingualift,fan2025slam} & Common Sense & \langbox{zh}, \langbox{en}, \langbox{de}, \langbox{fr}, \langbox{es}, \langbox{ru}, \langbox{hi} \\ \midrule
XNLI & \citep{zhu2024power,liu2024translation,zhang2024lingualift} & Logical & \langbox{zh}, \langbox{th}, \langbox{ur}, \langbox{en}, \langbox{de}, \langbox{fr}, \langbox{es}, \langbox{ru}, \langbox{el}, \langbox{tr}, \langbox{bg}, \langbox{hi}, \langbox{sw} \\ \midrule
MultiNLI & \citep{zhu2024power}, \citep{huang2024mindmerger} & Logical & \langbox{en} \\ \midrule
BBH-Hard & \citep{luo2024python} & Temporal, Tabular, Spatial & \langbox{Python}, \langbox{R}, \langbox{C++.} \langbox{Java}, \langbox{Javascript} \\ \midrule
NLVR2 & \citep{song2024missing} & Visual & \langbox{en} \\ \midrule
MARVL & \citep{song2024missing} & Visual & \langbox{id}, \langbox{sw}, \langbox{ta}, \langbox{tr}, \langbox{zh} \\ \midrule
xSTREET & \citep{li2024eliciting} & Logical & \langbox{ar}, \langbox{zh}, \langbox{ja}, \langbox{en}, \langbox{es}, \langbox{ru} \\ \midrule
Translated Code Comments (TCC) & \citep{li2024eliciting} & Code & \langbox{Java}, \langbox{JavaScript}, \langbox{Python} \\ \midrule
mCoT-MATH & \citep{lai2024mcot} & Maths & \langbox{zh}, \langbox{th}, \langbox{ja}, \langbox{en}, \langbox{de}, \langbox{fr}, \langbox{es}, \langbox{ru}, \langbox{bn}, \langbox{hi}, \langbox{te} \\ \midrule
Reasoning by Equivalence Dataset & \citep{arora2024towards} & Logical & \langbox{en}, \langbox{fr}, \langbox{es}, \langbox{de}, \langbox{pt}, \langbox{hi} \\ \midrule
Reasoning by Inheritance Dataset & \citep{arora2024towards} & Logical & \langbox{en}, \langbox{fr}, \langbox{es}, \langbox{de}, \langbox{pt}, \langbox{hi} \\ \midrule
XCOT & \citep{chai2401xcot} & Maths & \langbox{de}, \langbox{fr}, \langbox{es}, \langbox{ru}, \langbox{zh}, \langbox{ja}, \langbox{th}, \langbox{te}, \langbox{bn}, \langbox{sw}, \langbox{en} \\ \midrule
mCSQA & \citep{sakai2024mcsqa} & Common Sense & \langbox{zh}, \langbox{ja}, \langbox{en}, \langbox{fr}, \langbox{de}, \langbox{pt}, \langbox{ru} \\ \midrule
\end{tabularx}
\end{table}

\begin{table}[ht]
\centering
\small
\renewcommand{\arraystretch}{0.9}
\setlength{\tabcolsep}{1pt}  
\begin{tabularx}{\textwidth}{L{3cm} L{6cm} L{3cm} L{3cm}}
\toprule
\textbf{Dataset} & \textbf{Paper} & \textbf{Domain} & \textbf{Languages} \\
\midrule
Rulings, Legislation, Court View Generation, Critically Prediction, Law Area Prediction, Judgment Prediction Datasets & \citep{rasiah2024one} & Legal & \langbox{de}, \langbox{fr}, \langbox{it}, \langbox{ro}, \langbox{en} \\ \midrule
mRewardBench & \citep{gureja2024m} & Logical and CommonSense & \langbox{ar}, \langbox{cs}, \langbox{de}, \langbox{el}, \langbox{es}, \langbox{fa}, \langbox{fr}, \langbox{he}, \langbox{hi}, \langbox{id}, \langbox{it}, \langbox{ja}, \langbox{ko}, \langbox{nl}, \langbox{pl}, \langbox{pt}, \langbox{ro}, \langbox{ru}, \langbox{tr}, \langbox{uk}, \langbox{vi}, \langbox{zh} \\ \midrule
Moral Judgement Dataset & \citep{khandelwal2024moral} & Moral & \langbox{en}, \langbox{zh}, \langbox{hi}, \langbox{ru}, \langbox{es}, \langbox{sw} \\ \midrule
MCR & \citep{zhao2024large} & Compositional & \langbox{ja}, \langbox{ko}, \langbox{fr} \\ \midrule
mTEMPREASON & \citep{bajpai2025multilingual} & Temporal & \langbox{ro}, \langbox{de}, \langbox{fr} \\ \midrule
XCOPA & \citep{liu2024translation} & Common Sense & \langbox{zh}, \langbox{it}, \langbox{vi}, \langbox{tr}, \langbox{id}, \langbox{sw}, \langbox{th}, \langbox{et}, \langbox{ta}, \langbox{ht}, \langbox{qu} \\ \midrule
mARC & \citep{kargaran2024mexa} & Common Sense & \langbox{zh}, \langbox{ja}, \langbox{en}, \langbox{de}, \langbox{fr}, \langbox{es} \\ \midrule
IndiMathQA & \citep{anand2025multilingual} & Maths & \langbox{en}, \langbox{hi} \\ \midrule
CRUXEval & \citep{xu2024cruxeval} & Code & \langbox{C\#}, \langbox{C++}, \langbox{D}, \langbox{GO}, \langbox{Java}, \langbox{JavaScript}, \langbox{Julia}, \langbox{Luca}, \langbox{Perlm} \langbox{PHP}, \langbox{R}, \langbox{Racket}, \langbox{Ruby}, \langbox{Rust}, \langbox{Scala}, \langbox{Shell}, \langbox{Swift}, \langbox{TypeScript} \\ \midrule
\end{tabularx}
\end{table}

\begin{table}[ht]
\centering
\small
\renewcommand{\arraystretch}{0.9}
\setlength{\tabcolsep}{1pt}
\begin{tabularx}{\textwidth}{L{3cm} L{6cm} L{3cm} L{3cm}}
\toprule
\textbf{Dataset} & \textbf{Paper} & \textbf{Domain} & \textbf{Languages} \\
\midrule
mMMLU & \citep{kargaran2024mexa} & Common Sense &
\langbox{ar}, \langbox{zh}, \langbox{vi}, \langbox{id}, \langbox{en}, \langbox{de}, \langbox{fr}, \langbox{it}, \langbox{nl}, \langbox{eu}, \langbox{es}, \langbox{pt}, \langbox{ca}, \langbox{da}, \langbox{ru}, \langbox{hr}, \langbox{hy}, \langbox{hu}, \langbox{ro}, \langbox{ne}, \langbox{kn}, \langbox{uk}, \langbox{sr}, \langbox{sv}, \langbox{mr}, \langbox{nb}, \langbox{ml}, \langbox{is}, \langbox{bn}, \langbox{hi}, \langbox{ta}, \langbox{te}, \langbox{gu} \\
\midrule
MMWP Benchmark & \citep{zhang2024lingualift} & Maths &
\langbox{af}, \langbox{ar}, \langbox{be}, \langbox{bn}, \langbox{eu}, \langbox{gu}, \langbox{ha}, \langbox{hi}, \langbox{hy}, \langbox{is}, \langbox{kn}, \langbox{lb}, \langbox{mk}, \langbox{ml}, \langbox{mr}, \langbox{ne}, \langbox{sk}, \langbox{sw}, \langbox{ta}, \langbox{te}, \langbox{th}, \langbox{bg}, \langbox{ca}, \langbox{cs}, \langbox{da}, \langbox{fi}, \langbox{hr}, \langbox{hu}, \langbox{id}, \langbox{ko}, \langbox{nb}, \langbox{pl}, \langbox{pt}, \langbox{ro}, \langbox{sl}, \langbox{sr}, \langbox{uk}, \langbox{vi}, \langbox{de}, \langbox{en}, \langbox{es}, \langbox{fr}, \langbox{it}, \langbox{ja}, \langbox{nl}, \langbox{ru}, \langbox{sv}, \langbox{zh} \\
\bottomrule
\end{tabularx}
\end{table}

\noindent\textbf{Distribution of papers covering different aspects of Reasoning} 

\begin{table}[h!]
\centering
\small
\renewcommand{\arraystretch}{0.9}
\setlength{\tabcolsep}{1pt}  
\begin{tabular}{>{\raggedright\arraybackslash}p{3cm} p{13cm}} 
\toprule
\textbf{Reasoning Type} & \textbf{Papers} \\
\midrule
Deductive & \citet{lai2024mcot}, \citet{chai2401xcot}, \citet{huang2024mindmerger}, \citet{zhang2024lingualift}, \citet{huang2024mindmerger}, \citet{fan2025slam}, \citet{payoungkhamdee2024empirical}, \citet{luo2024python}, \citet{song2024missing}, \citet{li2024eliciting}, \citet{arora2024towards}, \citet{rasiah2024one}, \citet{sakai2024mcsqa}, \citet{khandelwal2024moral}, \citet{kargaran2024mexa}, \citet{anand2025multilingual}, \citet{xu2024cruxeval}, \citet{she2024mapo}, \citet{zhu2024power}, \citet{li2024quantifying}, \citet{lim2024mufu}, \citet{bajpai2025multilingual}, \citet{li2024improving} \\ \midrule
Inductive & \citet{chai2401xcot}, \citet{huang2024mindmerger}, \citet{zhang2024lingualift}, \citet{huang2024mindmerger}, \citet{fan2025slam}, \citet{payoungkhamdee2024empirical}, \citet{luo2024python}, \citet{song2024missing}, \citet{li2024eliciting}, \citet{arora2024towards}, \citet{rasiah2024one}, \citet{sakai2024mcsqa}, \citet{khandelwal2024moral}, \citet{kargaran2024mexa}, \citet{anand2025multilingual}, \citet{xu2024cruxeval}, \citet{she2024mapo}, \citet{zhu2024power}, \citet{li2024quantifying}, \citet{lim2024mufu}, \citet{bajpai2025multilingual}, \citet{li2024improving}, \citet{wei2024mlake}, \citet{xie2024memla}, \citet{yang2024language}, \citet{geng2024not}, \citet{yang2025mr}, \citet{ko2025understand}, \citet{ruan2025layalign}, \citet{lu2024dictionary}, \citet{agrawal2024evaluating}, \citet{ranaldi2025improving}, \citet{ha2025pensez}, \citet{ranaldi2025natural}, \citet{ranaldi2025multilingual}, \citet{xuan2025mmlu}, \citet{yoon2024langbridge}, \citet{zhu2024question}, \citet{lai2024mcot}, \citet{chai2401xcot}, \citet{huang2024mindmerger}, \citet{zhang2024lingualift}, \citet{huang2024mindmerger}, \citet{fan2025slam}, \citet{payoungkhamdee2024empirical}, \citet{luo2024python}, \citet{song2024missing}, \citet{li2024eliciting}, \citet{arora2024towards}, \citet{rasiah2024one}, \citet{sakai2024mcsqa}, \citet{khandelwal2024moral}, \citet{kargaran2024mexa}, \citet{anand2025multilingual}, \citet{xu2024cruxeval}, \citet{she2024mapo}, \citet{zhu2024power}, \citet{li2024quantifying}, \citet{lim2024mufu}, \citet{bajpai2025multilingual}, \citet{li2024improving}, \citet{wei2024mlake}, \citet{xie2024memla}, \citet{yang2024language}, \citet{geng2024not}, \citet{yang2025mr}, \citet{ko2025understand}, \citet{ruan2025layalign}, \citet{lu2024dictionary}, \citet{agrawal2024evaluating}, \citet{ranaldi2025improving}, \citet{ha2025pensez}, \citet{ranaldi2025natural}, \citet{ranaldi2025multilingual} \\ \midrule
Abductive & \citet{huang2024mindmerger}, \citet{zhang2024lingualift} \\ \midrule
Analogical & \citet{zhang2024lingualift}, \citet{huang2024mindmerger}, \citet{fan2025slam}, \citet{payoungkhamdee2024empirical}, \citet{luo2024python}, \citet{song2024missing}, \citet{li2024eliciting}, \citet{arora2024towards}, \citet{rasiah2024one}, \citet{sakai2024mcsqa}, \citet{khandelwal2024moral}, \citet{kargaran2024mexa}, \citet{anand2025multilingual}, \citet{xu2024cruxeval}, \citet{she2024mapo}, \citet{zhu2024power}, \citet{li2024quantifying}, \citet{lim2024mufu}, \citet{bajpai2025multilingual}, \citet{li2024improving}, \citet{wei2024mlake}, \citet{xie2024memla}, \citet{yang2024language}, \citet{geng2024not}, \citet{yang2025mr}, \citet{ko2025understand}, \citet{ruan2025layalign}, \citet{lu2024dictionary}, \citet{agrawal2024evaluating}, \citet{ranaldi2025improving}, \citet{ha2025pensez}, \citet{ranaldi2025natural}, \citet{ranaldi2025multilingual} \\ \midrule
Commonsense & \citet{huang2024mindmerger}, \citet{fan2025slam}, \citet{payoungkhamdee2024empirical}, \citet{luo2024python}, \citet{song2024missing}, \citet{li2024eliciting}, \citet{arora2024towards}, \citet{rasiah2024one}, \citet{sakai2024mcsqa}, \citet{khandelwal2024moral}, \citet{kargaran2024mexa}, \citet{anand2025multilingual}, \citet{xu2024cruxeval}, \citet{she2024mapo}, \citet{zhu2024power}, \citet{li2024quantifying}, \citet{lim2024mufu}, \citet{bajpai2025multilingual}, \citet{li2024improving}, \citet{wei2024mlake}, \citet{xie2024memla} \\
\bottomrule
\end{tabular}
\caption{Categorization of Papers by Reasoning Type}
\label{tab:reasoning_types}
\end{table}

\end{document}